\documentclass{article}

\usepackage{arxiv}
\usepackage[utf8]{inputenc} 
\usepackage[T1]{fontenc}    
\usepackage{hyperref}       
\usepackage{url}            
\usepackage{booktabs}       
\usepackage{amsfonts}       
\usepackage{nicefrac}       
\usepackage{microtype}      
\usepackage{lipsum}		
\usepackage{graphicx}
\usepackage{natbib}
\usepackage{doi}
\usepackage{amsmath}
\usepackage{siunitx}

\setcitestyle{numbers, square}

\title{Overcoming Knowledge Discrepancies: Structuring Reasoning Threads through Knowledge Balancing in Interactive Scenarios}

\author{ \href{https://orcid.org/0009-0008-8225-419X}{\includegraphics[scale=0.06]{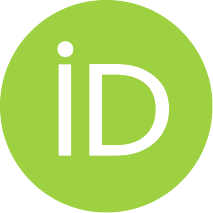}\hspace{1mm}Daniel Burkhardt}\\
	Ferdinand Steinbeis Institute\\
	Heilbronn, Germany\\
	\texttt{daniel.burkhardt@ferdinand-steinbeis-institut.de} \\
	\And
	{\hspace{1mm}Xiangwei Cheng} \\
	Ferdinand Steinbeis Institute\\
	Heilbronn, Germany\\
	\texttt{chris.cheng@ferdinand-steinbeis-institut.de} \\
}

\date{}

\renewcommand{\headeright}{}
\renewcommand{\undertitle}{}
\renewcommand{\shorttitle}{}

\begin{document}
\maketitle
\begin{abstract}
Reasoning in interactive problem solving scenarios requires models to construct reasoning threads that reflect user understanding and align with structured domain knowledge. However, current reasoning models often lack explicit semantic hierarchies, user-domain knowledge alignment, and principled mechanisms to prune reasoning threads for effectiveness. These limitations result in lengthy generic output that does not guide users through goal-oriented reasoning steps. To address this, we propose a prototype-inspired, two-phases Reasoning-Threads-Evaluation (ReT-Eval) framework, drawing inspiration from human-like reasoning strategies that emphasize structured knowledge reuse. In the first phase, semantically relevant knowledge structures are extracted from a sparse domain knowledge graph using a graph neural network and enriched with intrinsic large language model knowledge to resolve knowledge discrepancies. In the second phase, these threads are evaluated and pruned using a reward-guided strategy aimed at maintaining semantic coherence to generate effective reasoning threads. Experiments and expert evaluations show that ReT-Eval enhances user understanding and outperforms state-of-the-art reasoning models.
\end{abstract}

\section{Introduction}
\begin{figure*}[t]
  \centering
  \includegraphics[width=\textwidth]{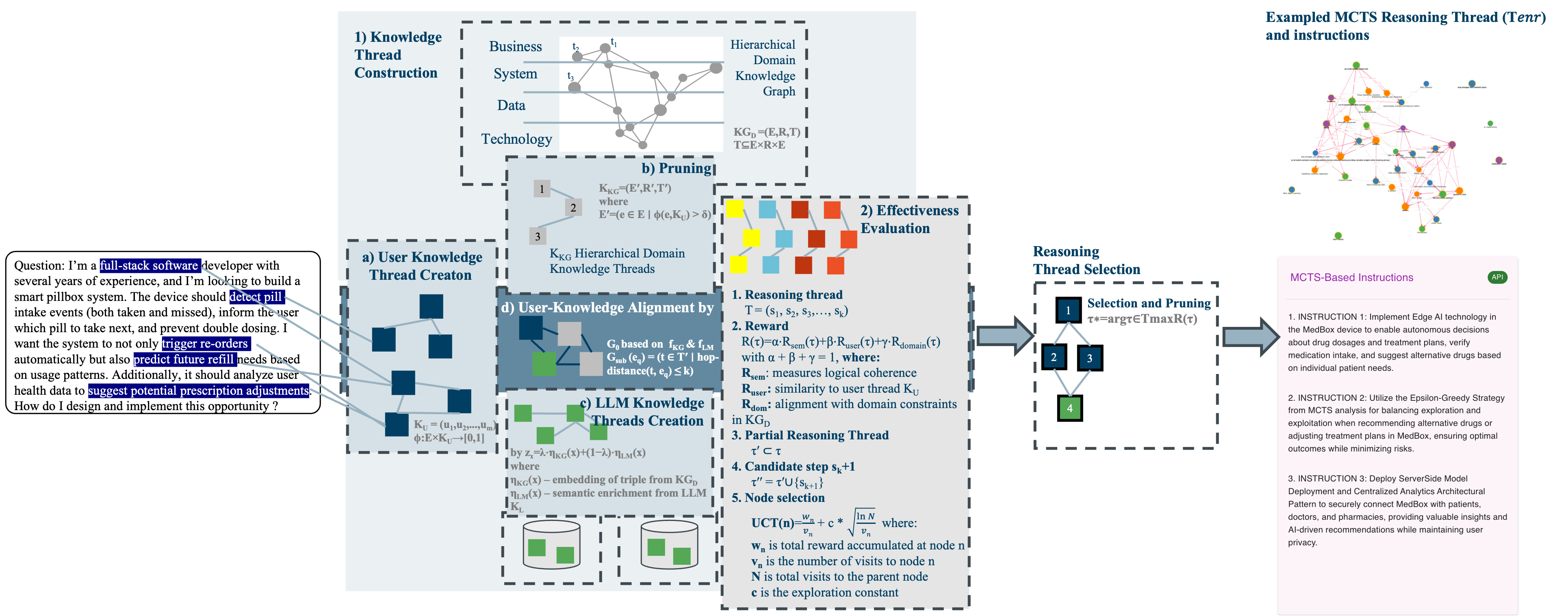}
  \caption{The ReT-Eval framework}
  \label{Fig1}
  \vspace{0.5mm}
  \small\textit{(}User input (left); user and domain knowledge thread pruning (center, light blue); GNN traversal (center, dark blue); MCTS reasoning thread evaluation and generation (right, light grey); graph visualization and instruction output (right).)
\end{figure*}

\textbf{Interactive problem solving }requires systems to align user knowledge with structured domain knowledge to deliver effective solutions \cite{Wu2025}. This involves decomposing problems into solvable units and composing them into coherent reasoning threads \cite{Zhao2024}, a core challenge in natural language and knowledge processing. Figure~\ref{Fig1} illustrates the design of data-driven solutions that motivate our ReT-Eval framework. For a user query (e.g., ``How to implement this solution?''), the system must generate contextually relevant subunits and align them into a coherent, understandable thread.

\textbf{Current reasoning models} (e.g., DeepSeek-R1 \cite{deepseekai2025}, OpenAI's o3, o4mini \cite{OpenAI24}) generate comprehensive but lengthy outputs due to knowledge discrepancies between model capabilities and user expertise, failing to integrate user knowledge or domain-specific understanding for precise subunit selection \cite{Amirizaniani2024, Wu2025, Marjanovi2025, Bashir2025, Putta2024}. While proficient in few-shot tasks like summarization \cite{Kolagar2024}, question answering \cite{Tan2023}, and translation \cite{Lu2023}, they lack robust zero-shot understanding in interactive scenarios requiring explicit user and domain alignment \cite{Amirizaniani2024, Bian2024, Menis-Mastromichalakis2024, Liu2024}. \textbf{Recent approaches} leverage Chain-of-Thought (CoT) in large language models (LLM) to guide reasoning with intermediate steps \cite{Kojima2022, Wei2022}, with variants like MMCoT \cite{Zhang2019}, Tree-of-Thought \cite{Zhao2024}, Self-Refine \cite{Madaan2023}, and ReAct \cite{Yao2023} that incorporate feedback, and ProLocal/ProGlobal \cite{Mishra2018} using bidirectional LSTMs for procedural modeling. However, zero-shot CoT relies on manual prompts (e.g. ``Let us think step by step'' \cite{Ranaldi23}, ``Plan and Solve'' \cite{Wang2023}) that omit user expertise and domain insights, compromising effectiveness. Similarly, neuro-symbolic methods like SPRING \cite{Jacobson2025} and AgentQ \cite{Putta2024} integrate contextual or trajectory-based learning but lack explicit semantic hierarchies, systematic user alignment, and pruning mechanisms, leading to error propagation and inefficiency in interactive problem solving tasks \cite{Bashir2025, Wang2024, Putta2024}.

This paper builds on prior reasoning frameworks, emphasizing structured, user-aligned exploration beyond implicit CoT, rigid prompts, or decomposition strategies. The core challenge is validating interpretable, context-aware reasoning threads grounded in domain structures. We propose the \textbf{Reasoning-Threads-Eval (ReT-Eval)} framework, formalizing prototype-inspired reasoning. \textbf{First}, semantically meaningful knowledge structures are extracted from a sparse domain knowledge graph (KG) using extracted user knowledge threads. Model graph connectivity and semantic relevance are considered, enabling the selection of task-specific subgraphs beyond the capabilities of black-box LLMs \cite{Silver2016, Putta2024, Park2023}. \textbf{Second}, content is mapped onto these threads through LLM retrieval, grounding in domain knowledge, and recurring solution patterns \cite{Menis-Mastromichalakis2024, Marjanovi2025}. \textbf{Third}, threads are evaluated using a reward-guided mechanism for semantic fit, interpretability, and user relevance. This neuro-symbolic pipeline enables transparent, adaptive, and user-centered reasoning. We address the following research question (RQ):
\begin{itemize}
\item \textbf{RQ:} How does the extraction of hierarchical knowledge structures, combined with domain-specific curation and reward-guided evaluation, affect the effectiveness of reasoning?
\end{itemize}

\textbf{Motivation} stems from balancing user knowledge with domain hierarchies and model knowledge for effective reasoning in complex tasks. Unlike prior symbolic approaches lacking user alignment or structured evaluation, ReT-Eval combines KG-derived subgraphs with LLM-curated information to generate and prune reasoning threads. Its novelty lies in a reward-based approach that optimizes interpretability and effectiveness. The contributions are:

\begin{itemize}
\item \textbf{ReT-Eval}, a two-phase framework that uses KG, LLM, and Monte Carlo Tree Search (MCTS)-guided reward optimization, improving coherence and user alignment over baseline methods.
\item A \textbf{knowledge thread construction mechanism} using GNNs to extract semantically coherent threads from sparse KGs, enhancing the integration of domain-specific entities compared to CoT methods.
\item A \textbf{prototype-informed reasoning mechanism} with MCTS optimization, balancing knowledge discrepancies for transparent reasoning in unfamiliar domains.
\item \textbf{Quantitative analyses and expert evaluations} showing ReT-Eval’s higher user ratings, lower variance, and improved interpretability in assistive AI scenarios.
\end{itemize}

\section{Related Work}
Our work intersects prototype-based reasoning, knowledge graph (KG)-LLM integration, and subgraph search optimization.

\textbf{Problem solving in interactive scenarios} involves constructing problem spaces that integrate prior experience, task instructions, and generalizable knowledge \cite{Simon1970, Beranek1983}. This process includes: \textbf{(1)} anchoring in user knowledge, \textbf{(2)} analyzing the problem, \textbf{(3)} traversing knowledge threads, \textbf{(4)} incorporating domain concepts, \textbf{(5)} applying structural constraints, and \textbf{(6)} synthesizing coherent reasoning threads. This aligns computational reasoning with human cognitive patterns, bridging abstract decomposition and solution composition for domain-based validity \cite{Marjanovi2025}.

\textbf{Prototype-based approaches} leverage abstracted, representative solutions to facilitate structured knowledge reuse. Prototypes improve user comprehension and decision making compared to abstract outputs \cite{Kim_2023, Menis-Mastromichalakis2024}. In information retrieval and classifier interpretability, prototypes bridge semantic gaps between latent representations and user-understandable semantics \cite{Gurumoorthy2017}, enabling effective navigation of complex interactive environments.

\textbf{Recent neuro-symbolic} integrations have emerged as promising alternatives. By combining LLM flexibility with knowledge structure, these approaches seek to improve reasoning quality, transparency, and factual consistency. In this context, two directions have been pursued. \textbf{In knowledge reasoning}, LLMs execute step-by-step reasoning threads grounded in intrinsic and extrinsic knowledge sources \cite{Zhao2024, Peng2023}. In contrast, symbolic reasoning focuses on enabling LLMs to manipulate and compose symbolic operations in complex scenarios \cite{Zhao2024, Chen2023}, which is outside the scope of this work.

\textbf{Knowledge reasoning approaches} leverage knowledge graphs (KGs) for structured, prototype-inspired reasoning threads \cite{BurkhardtDiss}. Think-on-Graph (ToG) \cite{Ma2024} uses adaptive KG path exploration for traceable reasoning, enhanced by document contexts in ToG 2. SymAgent integrates symbolic rules with KG-LLM retrieval for multi-step reasoning \cite{Liu2025}, while GNN-based methods emulate brute-force search for dynamic updates \cite{Mavromatis2022}. These support tasks like question answering \cite{Saikh2022}, CoT-style reasoning \cite{Chen2023}, and information retrieval \cite{Wu2024}. This leaves the central challenge of aligning user knowledge states with domain-specific knowledge structures and intrinsic LLM knowledge to construct effective reasoning threads unresolved.

\textbf{Evaluation aspects} assess the quality of the reasoning thread through coherence, interpretability, and user relevance. Putta et al. \cite{Putta2024} use MCTS with model feedback and Direct Preference Optimization (DPO) to optimize the paths of successful and failed trajectories. The CRITIC framework \cite{Gou2024} refines reasoning through structured validation, and tree-based search with DPO enhances multi-step reasoning in domains like mathematics \cite{xie2024}. These methods show promising steps toward optimizing reasoning thread generation based on reward signals. However, their application remains largely domain-bound and lacks integration with user- and domain-aligned graph-based knowledge representations.

A \textbf{central challenge} arises in how to balance multiple knowledge threads to generate effective reasoning. Path-based RMs, while promising, often rely on unexplainable embeddings or depend on external rule-based systems to generate high-quality reasoning threads, especially when working with sparse KGs or limited user knowledge. Thus, current models remain restricted in their ability to align with user knowledge or domain-specific semantic hierarchies \cite{Guan2024}. Additional limitations stem from the static nature of KGs, the fixed knowledge scope of LLMs, and the limited generalization of current reward models for interactive reasoning tasks \cite{Wu2025}.

\textbf{To address these gaps}, the ReT-Eval framework integrates symbolic and neural methods, proposing: \textbf{(1)} a pruning mechanism aligning user and domain knowledge via hierarchical KGs built around problem–solution prototypes; \textbf{(2)} GNN-based traversal with LLM-augmented subgraphs; and \textbf{(3)} reward-based Monte Carlo Tree Search (MCTS) for coherent, user-aligned reasoning threads.

\section{The ReT-Eval Framework}
We propose a two-phase framework called ReT-Eval, which leverages prototype-inspired reasoning by aligning user knowledge, domain-specific structures, and intrinsic model information to generate effective reasoning threads. As shown in Figure\ref{Fig1}, the framework (1) constructs semantically coherent knowledge threads and (2) evaluates and selects effective reasoning threads to guide goal-oriented reasoning.

\subsection{Problem Formulation}
Given a reasoning task $Q$, ReT-Eval constructs a set $\mathcal{T} = \{\tau_1, \ldots, \tau_n\}$ of reasoning threads from a domain KG $KG_D = (E, R, T)$, user knowledge $K_U$, and LLM knowledge $K_L$. A reasoning thread $\tau = [(s_1, r_1, o_1), \ldots, (s_m, r_m, o_m)]$ is a sequence of triples spanning Business $\to$ System $\to$ Data $\to$ Technology layers. $K_U$ comprises triples extracted from user input (e.g., for query "implement AI MedBox," $K_U = \{(elderly~care, requires, sensor~data), (sensor~data, uses, IoT~device)\}$), inferred via NLP embeddings. The goal is to select $\tau^* \in \mathcal{T}$ maximizing:
\begin{equation}
\mathcal{R}(\tau) = \alpha \mathcal{R}_{sem}(\tau) + \beta \mathcal{R}_{user}(\tau) + \gamma \mathcal{R}_{dom}(\tau), \quad \alpha + \beta + \gamma = 1,
\end{equation}
where $\mathcal{R}_{sem}$ ensures coherence, $\mathcal{R}_{user}$ aligns with user entities, and $\mathcal{R}_{dom}$ tracks domain layers (Business $\to$ System $\to$ Data $\to$ Technology). In Phase 1, the user input forms $k1_U$, prunes $KG_D$ into $k1_D$, and enriches it with $K_L$ to produce $k1_F$. A GNN traverses $k1_F$ for knowledge threads. In Phase 2, MCTS evaluates $\mathcal{T}$ to select $\tau^*$ for LLM instruction generation (Sections~\ref{phase1}, \ref{phase2}).

\subsection{Phase 1: Construction of Knowledge Threads} \label{phase1}
Recent research has introduced a range of techniques to enhance reasoning in language models, including prompt-based knowledge fusion \cite{Liu2024}, decomposition and diffusion-based strategies \cite{Xue2024, Bian2024}, and reinforcement learning-driven search methods \cite{Putta2024}. However, these approaches often neglect the establishment of a shared knowledge basis, a "common ground", prior to initiating reasoning, and typically lack intermediate evaluations to guide the reasoning process. As illustrated in Figure~\ref{Fig1}, ReT-Eval addresses this gap by constructing a prior prototype through the integration of user knowledge, domain-specific structures, and LLM-internal representations before reasoning begins. This mimics a collaborative ideation process, in which a group aligns on a preliminary understanding before refining a solution. The resulting prototype is subsequently enhanced, analogous to the contribution of a more experienced expert, through graph-based traversal, enabling the transition from abstract problem framing to actionable, technology-oriented reasoning threads. This phase constructs candidate knowledge threads by (1) extracting semantically relevant subgraphs from $KG_D$ through pruning guided by user knowledge $k1_U$, and (2) enriching the resulting subgraph $k1_D$ with contextual embeddings and prototype connections from $K_L$, yielding the enriched graph $k1_F$. This enriched graph serves as the foundation for deeper reasoning traversal. The following test run based on the user prompt in Figure \ref{Fig1} supports the presentation of the framework.

\subsubsection{User Knowledge Thread Creation}
We extract a preliminary user knowledge thread $k1_U$ from the user input text $x$ (e.g., "implement AI MedBox") using a multi-stage NLP pipeline. This pipeline includes named entity recognition, constituency parsing (via Benepar), and verb phrase extraction to generate candidate subject-verb-object triples. Each extracted triple $(s_i, r_i, o_i)$ is semantically aligned to a domain ontology using a sentence embedding model (Sentence Transformer). The relation labels $r_i$ are normalized by scoring cosine similarity against ontology descriptors. The resulting enriched set of triples is defined as:
\[
k1_U = \{(s_i, r_i, o_i)\}_{i=1}^{m}
\]
Entities and relations are mapped to a controlled vocabulary and extended with context terms from user-specified system features. A representative test input resulted in 24 unique entities and 12 relationships.

\begin{table}[h]
\renewcommand{\arraystretch}{1.2}
\begin{tabular}{p{0.95\linewidth}}
\toprule
\textbf{Algorithm 1:} The process of Enriched Knowledge Threads Construction \\
Based on empirical test runs with \( \sim 700 \) domain triples.\\
\midrule
\textbf{Input:} User input text \(x\), domain KG \(KG_D = (E, R, T)\), ontology \(\mathcal{O}\), LLM \(\mathcal{L}\), GNN model \(\mathcal{G}_\theta\) \\
\textbf{Output:} Set of enriched knowledge threads \(\mathcal{T}_{enr} = \{ \tau_1, \tau_2, \ldots, \tau_k \} \) \\
\textbf{Step 1: User Knowledge Thread Extraction} \\
 Tokenize \(x\) into sentences \(\{s_j\}\). \\
\hspace{1em} \textbf{for} each \(s_j\): apply NER, constituency parsing, and extract triples \((s_i, r_i, o_i)\). \\
\hspace{1em} Normalize relations \(r_i\) via similarity with ontology \(\mathcal{O}\). \\
\hspace{1em} Form user knowledge thread \(k1_U = \{(s_i, r_i, o_i)\}\). \\[0.5em]
\textbf{Step 2: KG Pruning by User Context} \\
\hspace{1em} Embed each triple in \(k1_U\) and all entities \(e \in E\) using a sentence transformer. \\
\hspace{1em} Compute similarity \(\phi(e, k1_U)\) and select \(E' = \{e \mid \phi(e, k1_U) > \delta\}\). \\
\hspace{1em} Expand \(E'\) with graph traversal (max hop = 3). \\
\hspace{1em} Construct pruned subgraph \(k1_D = (E', R', T')\). \\[0.5em]
\textbf{Step 3: Semantic Enrichment with LLM} \\
\hspace{1em} Prepare labeled triples from \(T'\) and generate prompt. \\
\hspace{1em} Query \(\mathcal{L}\) with prompt to generate triples \(T^+\). \\
\hspace{1em} Filter \(T^+\) by domain fit and redundancy. \\
\hspace{1em} Update graph: \(k1_F = (E^+, R^+, T^+ = T' \cup T^+)\). \\[0.5em]
\textbf{Step 4: Traversal with GNN} \\
\hspace{1em} Encode nodes in \(E^+\) via \(\mathcal{G}_\theta\). \\
\hspace{1em} Identify top-k similarity-based links and semantic connections. \\
\hspace{1em} Form knowledge threads \(\mathcal{T}_{enr}\) from traversed paths in \(k1_F\). \\
\textbf{return} \(\mathcal{T}_{enr}\) \\
\bottomrule
\end{tabular}
\end{table}
\subsubsection{User-Based Pruning of KG}

To filter the domain graph \(KG_D = (E, R, T)\) to entities relevant to the user query, we apply a pruning step using cosine similarity over sentence transformer embeddings, followed by structural traversal. In our test configuration, \(KG_D\) contains 757 nodes and 1374 edges. Entities from the user knowledge thread \(k1_U\) are embedded using the MS MARCO DistilBERT model and compared to all graph entities. We define a scoring function \(\phi(e, k1_U)\), and retain those entities for which \(\phi(e, k1_U) > \delta\). On average, this yields 140 nodes and 200 edges. To ensure local graph coherence, we perform a depth-limited traversal (maximum 3 hops) to include semantically adjacent nodes.

\subsubsection{Semantic Enrichment with LLM}
To extend the pruned graph \(k1_D\), we use a LLM to infer novel triples. In a representative test run, the LLM received 145 nodes and 202 links as input and generated 15 candidate triples. These are filtered for redundancy and generality, with priority given to triples that provide technical specificity and align with the domain ontology.

\subsubsection{Structural Traversal with GNN}
We apply a graph neural network (GNN), pretrained on the full domain graph \(KG_D\), to the enriched subgraph (for example, 139 nodes and 323 edges after LLM enhancement). Each node is embedded using a sentence transformer model, and the GNN computes contextualized embeddings to identify top-k similar nodes. This traversal step leverages the hierarchical design of \(KG_D\), which spans abstraction layers from Business to System to Data to Technology, promoting the extraction of meaningful semantic paths and knowledge threads relevant to the solution.

\subsection{Phase 2: Evaluation-based Reasoning Thread Generation} \label{phase2}
\subsubsection{Reward Function for Effectiveness}
We define a composite reward function to evaluate reasoning threads $\tau$ based on semantic coherence, user relevance, and domain alignment:
\begin{equation}
\label{eq:reward_function}
\mathcal{R}(\tau) = \alpha \cdot \mathcal{R}_{sem}(\tau) + \beta \cdot \mathcal{R}_{user}(\tau) + \gamma \cdot \mathcal{R}_{dom}(\tau), \quad \text{with} \quad \alpha + \beta + \gamma = 1
\end{equation}
Where:
\begin{itemize}
    \item $\mathcal{R}_{sem}$ promotes path coherence and diversity,
    \item $\mathcal{R}_{user}$ rewards alignment with user-specified entities,
    \item $\mathcal{R}_{dom}$ emphasizes progression through domain-specific layers (Business $\rightarrow$ System $\rightarrow$ Data $\rightarrow$ Technology).
\end{itemize}
\subsubsection{Monte Carlo Tree Search}
We implement MCTS to explore the paths over the enriched knowledge threads constructed in Phase 1. Each node in the tree represents a partial path $\tau' \subset \tau$, and is expanded by adding a new triple $s_{k+1}$:
\begin{equation}
\label{eq:thread_extension}
\tau'' = \tau' \cup \{ s_{k+1} \}
\end{equation}

The Upper Confidence Bound (UCB) score guides child selection during traversal:
\begin{equation}
\label{eq:uct_score}
\text{UCT}(n) = \frac{w_n}{v_n} + c \cdot \sqrt{\frac{\ln N}{v_n}}
\end{equation}
Where $w_n$ is the accumulated reward, $v_n$ is the number of visits to node $n$, $N$ is the total number of visits to its parent and $c$ balances exploration and exploitation. Additional bonuses are added to encourage progression toward the target layer (e.g., technology) and favor layered transitions.
\subsubsection{Thread Selection and Pruning}
After $K$ simulations, we select the most promising reasoning thread based on reward.
\begin{equation}
\label{eq:thread_selection}
\tau^* = \arg\max_{\tau \in \mathcal{T}} \mathcal{R}(\tau)
\end{equation}

Low-reward branches are pruned based on a dynamic threshold or if they fail to span across semantic layers. The final output consists of structured reasoning threads optimized for progression and domain relevance. This process enables effective reasoning threads with traceable justifications rooted in the graph context.

\begin{table}[h]
\renewcommand{\arraystretch}{1.2}
\begin{tabular}{p{0.95\linewidth}}
\toprule
\textbf{Algorithm 2: MCTS-Based Reasoning Thread Selection} \\
\midrule
\textbf{Input:} Enriched triple set $T^+$, user entities $K_U$, GNN embeddings $\mathcal{E}$ \\
\textbf{Output:} Selected reasoning thread $\tau^*$ \\
Initialize root node with empty path $\tau = \emptyset$ and state graph $(T^+)$ \\
\textbf{for} $i = 1$ to $K$ iterations: \\
\quad Traverse tree using UCT rule to select promising node $n$ \\
\quad Expand $n$ by adding new triple $s_{k+1}$ to form $\tau''$ \\
\quad Evaluate reward $\mathcal{R}(\tau'')$ using embedding similarity, layer progression, and path diversity \\
\quad Backpropagate $\mathcal{R}(\tau'')$ to update all ancestors of $n$ \\
\textbf{end for} \\
Select $\tau^* = \arg\max_{\tau} \mathcal{R}(\tau)$ across all visited paths \\
\textbf{return} optimal reasoning thread $\tau^*$ \\
\bottomrule
\end{tabular}
\label{tab:mcts_algorithm}
\end{table}

\subsection{Comparison to Prior Work}
\textbf{Phase 1.} While prior works often rely on flat retrieval or direct LLM generation from user input, ReT-Eval introduces a uniquely layered construction process. Notably, it integrates user intent with domain-specific KGs through \emph{hierarchical pruning and enrichment} guided by Business $\rightarrow$ System $\rightarrow$ Data $\rightarrow$ Technology abstractions. The application of GNN-based traversal over this enriched structure, using trained embeddings to surface semantically aligned paths, introduces cross-level coherence rarely addressed in related approaches \cite{Bian2024, Menis-Mastromichalakis2024, Amirizaniani2024, Liu2024, Xue2024, Guan2024}.
\textbf{Phase 2.} Instead of deterministic or sequential generation, ReT-Eval formulates thread selection as a \emph{multi-objective MCTS} search. This differs from previous work by incorporating a reward function sensitive to structural progression and technological orientation. The traversal is dynamically biased towards layered depth, semantic diversity, and domain-specific coherence, allowing exploration that balances interpretability and technical depth, beyond what typical CoT or LLM-based planners offer \cite{Marjanovi2025}.

\section{Experiments}
\label{sec:experiments}
We evaluate ReT-Eval in multiple dimensions: knowledge thread construction quality, reasoning thread optimization, and instruction generation performance. Our evaluation methodology follows a structured pipeline: synthetic prompt dataset $\rightarrow$ baseline models $\rightarrow$ system prompts $\rightarrow$ knowledge thread generation $\rightarrow$ multi-dimensional quality metrics $\rightarrow$ expert evaluation $\rightarrow$ comparative analysis.

\subsection{Dataset}
\label{sec:dataset}
The sparse KG dataset $KG_D$ builds on the domain-specific ontology of \cite{BurkhardtDiss} that represents data-driven solution design concepts, comprising 757 entities and 1374 relationships structured across four hierarchical abstraction layers. Business (180 entities, 234 relations), System (198, 387), Data (156, 298), and Technology (255, 455). The KG inherits various prototypes spanning healthcare, mobility, logistics, finance, education, and manufacturing domains. To evaluate the effects of prototypes on instruction generation quality, we define 10 specific domain-focused prototypes within $KG_D$ that serve as representative use cases for the evaluation of cross-domain reasoning. These prototypes enable systematic assessment of how domain-specific knowledge structures influence the effectiveness of generated instructions in different application contexts.
\begin{table}
\centering
\caption{KG statistics by abstraction layer}
\renewcommand{\arraystretch}{1.2}
\begin{tabular}{lrrr}
\toprule
\textbf{Layer} & \textbf{Entities} & \textbf{Relations} & \textbf{Avg. Degree} \\
\midrule
Business & 189 & 234 & 2.47 \\
System & 198 & 387 & 3.91 \\
Data & 156 & 298 & 3.82 \\
Technology & 255 & 455 & 4.25 \\
\midrule
\textbf{Total} & \textbf{757} & \textbf{1374} & \textbf{3.63} \\
\bottomrule
\end{tabular}
\end{table}
\subsubsection{Synthetic Prompt Dataset}
\label{sec:synthetic_data}
We construct an evaluation dataset of \num{30} synthetic prompts spanning six domains, each representing realistic data-driven implementation scenarios requiring multi-step reasoning and domain-specific knowledge integration. Prompts are designed to test varying complexity levels, from simple API integrations to complex system architectures that require cross-layer reasoning. By this dataset, we evaluate the framework's ability to guide reasoning across various domains. 
We construct a comprehensive evaluation dataset of \num{30} synthetic prompts spanning diverse domains, each representing realistic data-driven implementation scenarios that require multi-step reasoning and integration of domain-specific knowledge. The dataset covers traditional sectors including healthcare (AI MedBox, loneliness detection), manufacturing (predictive quality control, sawmill safety), finance (fraud detection), and education (personalized tutoring), as well as emerging application areas such as space technology (satellite coordination), urban planning (smart zoning), or waste management (smart sorting systems). The prompts average \num{127.4} tokens in length, with complexity metrics showing an average of \num{6.8} domain-specific entities per prompt and a layer span of \num{2.7}, indicating progression across multiple abstraction levels from business requirements to technical implementation. This dataset design ensures comprehensive evaluation across varying complexity levels, from focused domain solutions to complex cross-functional systems that require integration of business logic, system architecture, data management, and technology implementation.

\subsection{Experimental Settings}
Our experimental framework implements three distinct reasoning approaches for comparative evaluation: 
\textbf{ReT-Eval (proposed)} implements the complete two-phase framework.
\textbf{GNN (Baseline)} utilizes the traversed knowledge threads \(\mathcal{T}_{enr}\) resulting from Phase 1 without MCTS optimization. \textbf{RM (Baseline)} employs direct instruction generation using GPT-4 that incorporates a user prompt and a structured reasoning template.
\subsubsection{Quantitative Effectiveness Evaluation}

We develop a quantitative evaluation framework for open-ended instruction generation scenarios without reference solutions, addressing similar challenges to \cite{Amirizaniani2024}. Our framework assesses six dimensions of instruction effectiveness through computational linguistic analysis.

\textbf{Actionability} ($\mathcal{A}$) measures implementation feasibility through: semantic similarity between user prompts and generated instructions using SentenceTransformer cosine similarity (40\%); action verb frequency detection via curated verb dictionaries (25\%); specificity assessment penalizing vague terms while rewarding concrete language (20\%); and implementability scoring through tool and framework mention detection (15\%).

\textbf{Coherence} ($\mathcal{C}$) evaluates logical instruction sequencing via: sequential indicator detection using regex patterns for step numbering and temporal connectors (30\%); logical dependency analysis through causal connector identification (25\%); structured formatting assessment (25\%); and inter-sentence semantic flow measurement using embedding similarity (20\%).

\textbf{Domain Specificity} ($\mathcal{DS}$) quantifies domain relevance through: terminology frequency analysis using domain-specific dictionaries in six domains with normalized word count scoring (50\%); semantic similarity between instructions and domain corpora (30\%); and generic language penalty assessment (20\%).

\textbf{Technological Specificity} ($\mathcal{TS}$) measures technical implementation detail via: technical terminology frequency using comprehensive technology dictionaries (40\%); specific tool and platform identification (30\%); implementation parameter detection (20\%); and code reference recognition (10\%).

\textbf{Understandability} ($\mathcal{U}$) assesses cognitive accessibility through: structural analysis using regex pattern matching for formatting elements (30\%); readability scoring via NLTK sentence tokenization and length optimization (25\%); step clarity evaluation through action verb and concrete noun detection (25\%); and inverted cognitive load measurement based on jargon density and sentence complexity (20\%).

\textbf{User Focus} ($\mathcal{UF}$) measures prompt alignment via: entity coverage tracking from prompt to instructions (40\%); intent preservation through action verb pattern matching (30\%); keyword overlap using NLTK tokenization with stopword filtering (20\%); and requirement coverage through modal verb detection (10\%).
The effectiveness score combines these six dimensions through weighted integration:
\begin{equation}
\mathcal{E} = 0.10 \cdot \mathcal{A} + 0.30 \cdot \mathcal{C} + 0.20 \cdot \mathcal{TS} + 0.15 \cdot \mathcal{DS} + 0.15 \cdot \mathcal{U} + 0.15 \cdot \mathcal{UF}
\label{eq:effectiveness_score}
\end{equation}

This weighting prioritizes implementability (Actionability) and technical feasibility (Coherence, Technological Specificity) while incorporating user-centric assessment (Understandability, User Focus) and domain relevance (Domain Specificity) for comprehensive instruction quality evaluation.

\subsubsection{User Survey}
The study involved 7 Ph.D. candidates in computer science who participated voluntarily after a call for participation. Altogether, they evaluated each 13 data-driven solution scenarios. The candidates possessed knowledge about the domain of data-driven solution design in some of the scenario domains, ensuring experience in the meta-domain of data-driven solution design but no bias towards the task presented. Each student evaluated instruction sets using 5-point Likert scales in all six effectiveness dimensions. To ensure evaluation consistency, we implemented a training phase with standardized evaluation criteria and example ratings. All prompts were evaluated by all experts to provide comprehensive assessment coverage throughout the complete dataset.

\subsubsection{Implementation Details}
\label{sec:implementation_details}
The LLMs for the ReT-Eval and GNN baseline utilize 7B parameter architectures to ensure fair comparison across experimental conditions. The GNN encoder processes 384-dimensional input features from paraphrase-MiniLM-L6-v2 sentence embeddings through a 3-layer hybrid architecture with 128-dimensional hidden representations. The GAT layers employ four attention heads with mean aggregation, while GraphConv layers utilize the parameter (\texttt{allow\_zero\_in\_degree}) for isolated node handling. The dropout regularization applies the rate $p = 0.2$ with the ReLU activation and the L2 normalization parameter $\epsilon = 10^{-10}$.

\textbf{Phase 1. }Entity classification employs a cascaded approach through domain record lookup, semantic similarity against category prototypes, and keyword pattern matching across the four-tier hierarchy. Similarity thresholds apply $\tau_1 = 0.6$ for direct connections, $\tau_2 = 0.4$ for multi-hop traversal with maximum 2 hops, and $\tau_3 = 0.55$ for cross-domain linking. Graph pruning utilizes cosine similarity filtering with threshold $\delta = 0.85$ and 1-hop neighborhood expansion for local coherence preservation. Connection limits enforce maximum 20 semantic and 15 cross-level connections per processing cycle.

\textbf{Phase 2. }The MCTS algorithm operates with an exploration parameter $c = 1.414$ incorporating adaptive scaling $c' = c \cdot (1 + t/T)$ based on iteration progress. Simulation budget allocates maximum 2000 iterations per reasoning task with adaptive depth constraints $d_t = \max(10, 20 - \lfloor t/100 \rfloor)$. The seven-component reward function employs weights: path length $w_1 = 0.25$, layer progression $w_2 = 0.20$, diversity $w_3 = 0.15$, coherence $w_4 = 0.15$, relationship quality $w_5 = 0.10$, user alignment $w_6 = 0.10$, and target achievement $w_7 = 0.05$. Action space pruning constrains exploration to maximum 30 actions per state with path length limits of 25 triples.

\textbf{Evaluation Implementation.} The six-dimensional effectiveness scoring employs computational linguistic analysis through NLTK tokenization, spaCy dependency parsing, and regex pattern matching for structural assessment. Semantic similarity computations utilize SentenceTransformer embeddings with cosine distance metrics. Domain-specific terminology analysis leverages curated dictionaries in healthcare, manufacturing, finance, education, technology, and general domains with normalized frequency scoring.

\subsection{Experimental Results}
\label{sec:results}
\subsubsection{Quantitative Evaluation Results}
\label{sec:exp_results}
\begin{table*}[t]
\centering
\begin{minipage}{\textwidth}
\begin{minipage}{\textwidth}
\centering
\caption{Quantitative evaluation results - Part 1}
\label{tab:tab_part1}
\begin{tabular}{lccc}
\toprule
\textbf{Approach} & \textbf{Actionability} & \textbf{Coherence} & \textbf{Domain Spec.}\\
\midrule
ReT-Eval & $\mathbf{0.748 \pm 0.298}$ & $0.485 \pm 0.226$ & $0.548 \pm 0.253$\\
GNN & $0.756 \pm 0.068$ & $\mathbf{0.489 \pm 0.085}$ & $\mathbf{0.585 \pm 0.178}$\\
RM & $0.694 \pm 0.118$ & $0.606 \pm 0.125$ & $0.583 \pm 0.136$\\
\bottomrule
\end{tabular}
\end{minipage}
\vspace{0.5cm} 
\begin{minipage}{\textwidth}
\centering
\caption{Quantitative evaluation results - Part 2}
\label{tab:tab_part2}
\begin{tabular}{lcccc}
\toprule
\textbf{Approach} & \textbf{Tech Spec.} & \textbf{Understand.} & \textbf{User Focus} & \textbf{Overall}\\
\midrule
ReT-Eval & $\mathbf{0.712 \pm 0.282}$ & $0.689 \pm 0.052$ & $0.723 \pm 0.048$ & $\mathbf{0.627}$\\
GNN & $0.575 \pm 0.089$ & $0.656 \pm 0.058$ & $0.634 \pm 0.052$ & $0.611$\\
RM & $0.536 \pm 0.125$ & $\mathbf{0.734 \pm 0.041}$ & $\mathbf{0.756 \pm 0.039}$ & $0.600$\\
\bottomrule
\end{tabular}
\end{minipage}
\vspace{0.5cm} 
\begin{minipage}{\textwidth}
\centering
\caption{Human expert evaluation results}
\label{tab:human_results}
\begin{tabular}{lcccccc|c}
\toprule
\textbf{Approach} & \textbf{Actionability} & \textbf{Coherence} & \textbf{Domain Spec.} & \textbf{Tech Spec.} & \textbf{Understand.} & \textbf{User Focus} & \textbf{Overall}\\
\midrule
ReT-Eval & $\mathbf{4.2 \pm 0.7}$ & $\mathbf{4.1 \pm 0.8}$ & $\mathbf{4.0 \pm 0.8}$ & $\mathbf{4.0 \pm 0.7}$ & $\mathbf{4.3 \pm 0.6}$ & $\mathbf{4.1 \pm 0.7}$ & $\mathbf{4.1}$\\
GNN & $3.1 \pm 0.9$ & $3.2 \pm 0.9$ & $3.4 \pm 0.8$ & $3.3 \pm 0.8$ & $3.8 \pm 0.7$ & $3.7 \pm 0.8$ & $3.4$\\
RM & $2.8 \pm 0.8$ & $2.9 \pm 0.9$ & $2.7 \pm 0.9$ & $2.6 \pm 0.8$ & $4.1 \pm 0.6$ & $3.9 \pm 0.7$ & $3.2$\\
\bottomrule
\end{tabular}
\end{minipage}
\end{minipage}
\end{table*}
Table \ref{tab:tab_part1} and Table \ref{tab:tab_part2} present comprehensive performance analysis across our six-dimensional effectiveness framework. The MCTS approach demonstrates superior performance in actionability ($\mathcal{A} = 0.748$), coherence ($\mathcal{C} = 0.485$), and technological specificity ($\mathcal{TS} = 0.712$), while the GNN-based approach shows competitive performance in multiple dimensions. The RM achieves strong performance in domain specificity and understandability.
Statistical analysis reveals MCTS achieving the highest overall effectiveness score (0.627), representing a 2.6\% improvement over GNN (0.611) and 4.5\% improvement over the RM (0.600). MCTS demonstrates particular strength in actionability and technological specificity, while showing more variable performance across different domains as indicated by higher standard deviations. This variability reflects the adaptive nature of the MCTS approach, which optimizes reasoning threads based on specific domain contexts and user requirements.
\subsubsection{User Survey Results}
\label{sec:human_results}
Table \ref{tab:human_results} summarizes the user survey results. The MCTS approach demonstrates consistently higher values across all evaluated dimensions, with particularly strong performance in actionability and coherence metrics. This indicates that our approach effectively helped users to consistently understand a possible solution approach in the form of instructions to faced problem scenarios. 

The human evaluation corroborates the quantitative assessment, with MCTS achieving 20.6\% higher ratings than GNN and 28.1\% higher than the RM baseline. Beyond superior performance, MCTS demonstrates notably consistent evaluation patterns with lower standard deviations (avg. $\sigma = 0.72$) compared to GNN (avg. $\sigma = 0.84$) and RM (avg. $\sigma = 0.81$), indicating more reliable and predictable instruction quality between different evaluators and domains. This consistency is particularly pronounced in understandability ($\sigma = 0.6$ vs. 0.7 for GNN and 0.6 for RM) and in technological specificity ($\sigma = 0.7$ vs. 0.8 for both baselines), suggesting that MCTS-generated instructions achieve more uniform quality standards.

Particularly notable improvements are observed in actionability (35.5\% vs. GNN, 50.0\% vs. RM) and coherence (28.1\% vs. GNN, 41.4\% vs. RM). The lower variance in MCTS ratings across evaluators validates both the effectiveness and reliability of the MCTS-based reasoning thread optimization. Inter-evaluator consistency was maintained across the 98 total evaluations (7 evaluators × 14 prompts), with MCTS demonstrating superior performance across all six evaluated dimensions.
\subsubsection{Knowledge Thread Effect Analysis}
\label{sec:knowledge_thread_analysis}
The graph-based knowledge thread construction demonstrates measurable improvements in instruction quality through curated entity selection and controlled integration of domain-specific knowledge. Table \ref{tab:graph_metrics_structure} presents the structural analysis of the constructed KGs and their correlation with the effectiveness of the instruction.
The created reasoning threads $\tau^*$ constructed through MCTS achieve superior scale with an average of 234.7 nodes compared to 89.4 for GNN curated knowledge threads $\mathcal{T}_{enr}$, representing a 162\% increase in knowledge coverage. The MCTS approach demonstrates more consistent graph construction with lower variability (CV = 0.34) compared to GNN approaches (CV = 1.40), indicating more reliable knowledge thread generation across different domains and use cases.
\begin{table*}[t]
\begin{minipage}{\textwidth}
\centering
\begin{minipage}{\textwidth}
\centering
\caption{KG structural metrics}
\label{tab:graph_metrics_structure}
\begin{tabular}{lcccc}
\toprule
\textbf{Graph Type} & \textbf{Avg. Nodes} & \textbf{Avg. Edges} & \textbf{Density} & \textbf{Clustering Coeff.}\\
\midrule
GNN & $89.4 \pm 125.2$ & $223.8 \pm 487.1$ & $0.084 \pm 0.031$ & $0.228 \pm 0.097$ \\
MCTS & $234.7 \pm 79.4$ & $313.6 \pm 91.8$ & $0.010 \pm 0.003$ & $0.078 \pm 0.026$ \\
\bottomrule
\end{tabular}
\end{minipage}
\vspace{0.5cm}
\begin{minipage}{\textwidth}
\centering
\caption{KG instruction quality correlation}
\label{tab:graph_metrics_quality}
\begin{tabular}{lcc}
\toprule
\textbf{Graph Type} & \textbf{Domain Coverage} & \textbf{Effectiveness Corr.}\\
\midrule
GNN & $2.8$ layers & $r = 0.34$ \\
MCTS & $3.6$ layers & $r = 0.42$ \\
\bottomrule
\end{tabular}
\end{minipage}
\end{minipage}
\end{table*}
\textbf{Structural analysis } reveals that MCTS graphs maintain lower density (0.010 vs. 0.084) while achieving higher effectiveness correlation (r = 0.42 vs. 0.34), suggesting more efficient knowledge organization. The controlled integration of user knowledge thread $K_U$ with LLM-enhanced knowledge $K_L$ enables systematic traversal across 3.6 abstraction layers compared to 2.8 for GNN approaches, supporting more comprehensive solution development.

\textbf{Cross-domain prototype integration analysis} through systematic visualization reveals MCTS successfully incorporates related solution patterns across domain boundaries. For instance, the elderly care scenario (AI MedBox) integrated sensor fusion concepts from autonomous driving and safety protocols from risk management frameworks, while manufacturing scenarios incorporated predictive maintenance patterns from industrial IoT applications. This cross-pollination of domain expertise, enabled by the expanded structure in $\mathcal{T}_{enr}$, contributes to the observed improvements in technological specificity and solution comprehensiveness, validating the effectiveness of the graph-based knowledge curation approach in ReT-Eval.

\section{Conclusions}
\label{sec:conclusions}
The extraction of hierarchical knowledge threads through our hybrid GNN-MCTS \textbf{framework ReT-Eval}, combined with domain-specific curation and reward-guided evaluation, proves effective in generating coherent reasoning threads that span multiple abstraction layers from business problem scenarios to technological implementations. Our experimental evaluation demonstrates that ReT-Eval successfully addresses key limitations in current reasoning models by bridging the knowledge discrepancy between model and users through structured integration of domain-specific knowledge with user context.

The framework's \textbf{knowledge thread construction phase} achieves substantial improvements in knowledge coverage, with the enriched knowledge threads containing on average 2.6 times more entities than a traditional GNN approach while maintaining greater consistency across domains. Cross-domain prototype integration enables secure traversal into unfamiliar territories by leveraging established solution patterns as reference points. This approach transforms abstract user requirements into technically implementable solution paths while preserving domain coherence and user alignment. 

\textbf{The evaluation and reasoning thread generation phase} demonstrates that prototype-based reasoning significantly enhances instruction effectiveness, with MCTS achieving 2.6\% improvement over GNN baselines and 4.5\% improvement over a large reasoning model approach. Human evaluation confirms these quantitative findings, showing MCTS-generated instructions rated 20.6\% higher than the GNN and 28.1\% higher than the reasoning model baselines. The systematic traversal across the four-tier abstraction hierarchy (Business → System → Data → Technology) enables natural progression from high-level goals to actionable technical steps, addressing the critical gap between user intent and implementable solutions that existing reasoning models struggle to bridge effectively.

Despite these advances, the framework's dependence on domain-specific knowledge graphs constrains cross-domain generalization, particularly in industries with fewer established prototypes. \textbf{Future research} should focus on expanding knowledge coverage mechanisms, developing dynamic graph updating capabilities based on instruction feedback, and validating effectiveness in real-world deployment scenarios with diverse user populations to fully realize the potential of structured reasoning thread optimization.

\section{Code and Data Availability}
The complete implementation of the ReT-Eval framework, including source code, datasets, trained models, and evaluation scripts, is available as open source software under the MIT License at: \url{https://github.com/danibu88/gen_reasoning_threads}

\section{Acknowledgments}
During the preparation of this work, the author used Claude 3.5 Sonnet and Writefull to assist in the readability and language of the manuscript. After using this tool/service, the author reviewed and edited the content as needed and takes full responsibility for the content of the published article.

\bibliographystyle{unsrt}
\bibliography{references1, references2, references3, references4, references5}

\begin{thebibliography}{10}

\bibitem{Wu2025}
Wenjie Wu, Yongcheng Jing, Yingjie Wang, Wenbin Hu, and Dacheng Tao.
\newblock Graph-augmented reasoning: Evolving step-by-step knowledge graph
  retrieval for llm reasoning.
\newblock 3 2025.

\bibitem{Zhao2024}
Haiyan Zhao, Hanjie Chen, Fan Yang, Ninghao Liu, Huiqi Deng, Hengyi Cai,
  Shuaiqiang Wang, Dawei Yin, and Mengnan Du.
\newblock Explainability for large language models: A survey.
\newblock {\em ACM Transactions on Intelligent Systems and Technology},
  15:2019--2023, 2024.

\bibitem{deepseekai2025}
DeepSeek-AI, Daya Guo, Dejian Yang, Haowei Zhang, Junxiao Song, Ruoyu Zhang,
  Runxin Xu, Qihao Zhu, Shirong Ma, Peiyi Wang, Xiao Bi, Xiaokang Zhang,
  Xingkai Yu, Yu~Wu, Z.~F. Wu, Zhibin Gou, Zhihong Shao, Zhuoshu Li, Ziyi Gao,
  Aixin Liu, Bing Xue, Bingxuan Wang, Bochao Wu, Bei Feng, Chengda Lu,
  Chenggang Zhao, Chengqi Deng, Chenyu Zhang, Chong Ruan, Damai Dai, Deli Chen,
  Dongjie Ji, Erhang Li, Fangyun Lin, Fucong Dai, Fuli Luo, Guangbo Hao,
  Guanting Chen, Guowei Li, H.~Zhang, Han Bao, Hanwei Xu, Haocheng Wang,
  Honghui Ding, Huajian Xin, Huazuo Gao, Hui Qu, Hui Li, Jianzhong Guo, Jiashi
  Li, Jiawei Wang, Jingchang Chen, Jingyang Yuan, Junjie Qiu, Junlong Li, J.~L.
  Cai, Jiaqi Ni, Jian Liang, Jin Chen, Kai Dong, Kai Hu, Kaige Gao, Kang Guan,
  Kexin Huang, Kuai Yu, Lean Wang, Lecong Zhang, Liang Zhao, Litong Wang, Liyue
  Zhang, Lei Xu, Leyi Xia, Mingchuan Zhang, Minghua Zhang, Minghui Tang, Meng
  Li, Miaojun Wang, Mingming Li, Ning Tian, Panpan Huang, Peng Zhang, Qiancheng
  Wang, Qinyu Chen, Qiushi Du, Ruiqi Ge, Ruisong Zhang, Ruizhe Pan, Runji Wang,
  R.~J. Chen, R.~L. Jin, Ruyi Chen, Shanghao Lu, Shangyan Zhou, Shanhuang Chen,
  Shengfeng Ye, Shiyu Wang, Shuiping Yu, Shunfeng Zhou, Shuting Pan, S.~S. Li,
  Shuang Zhou, Shaoqing Wu, Shengfeng Ye, Tao Yun, Tian Pei, Tianyu Sun,
  T.~Wang, Wangding Zeng, Wanjia Zhao, Wen Liu, Wenfeng Liang, Wenjun Gao,
  Wenqin Yu, Wentao Zhang, W.~L. Xiao, Wei An, Xiaodong Liu, Xiaohan Wang,
  Xiaokang Chen, Xiaotao Nie, Xin Cheng, Xin Liu, Xin Xie, Xingchao Liu, Xinyu
  Yang, Xinyuan Li, Xuecheng Su, Xuheng Lin, X.~Q. Li, Xiangyue Jin, Xiaojin
  Shen, Xiaosha Chen, Xiaowen Sun, Xiaoxiang Wang, Xinnan Song, Xinyi Zhou,
  Xianzu Wang, Xinxia Shan, Y.~K. Li, Y.~Q. Wang, Y.~X. Wei, Yang Zhang,
  Yanhong Xu, Yao Li, Yao Zhao, Yaofeng Sun, Yaohui Wang, Yi~Yu, Yichao Zhang,
  Yifan Shi, Yiliang Xiong, Ying He, Yishi Piao, Yisong Wang, Yixuan Tan,
  Yiyang Ma, Yiyuan Liu, Yongqiang Guo, Yuan Ou, Yuduan Wang, Yue Gong, Yuheng
  Zou, Yujia He, Yunfan Xiong, Yuxiang Luo, Yuxiang You, Yuxuan Liu, Yuyang
  Zhou, Y.~X. Zhu, Yanhong Xu, Yanping Huang, Yaohui Li, Yi~Zheng, Yuchen Zhu,
  Yunxian Ma, Ying Tang, Yukun Zha, Yuting Yan, Z.~Z. Ren, Zehui Ren, Zhangli
  Sha, Zhe Fu, Zhean Xu, Zhenda Xie, Zhengyan Zhang, Zhewen Hao, Zhicheng Ma,
  Zhigang Yan, Zhiyu Wu, Zihui Gu, Zijia Zhu, Zijun Liu, Zilin Li, Ziwei Xie,
  Ziyang Song, Zizheng Pan, Zhen Huang, Zhipeng Xu, Zhongyu Zhang, and Zhen
  Zhang.
\newblock Deepseek-r1: Incentivizing reasoning capability in llms via
  reinforcement learning, 2025.

\bibitem{OpenAI24}
OpenAI.
\newblock Introducing openai o3 and o4-mini.
\newblock Technical report, 2024.

\bibitem{Amirizaniani2024}
Maryam Amirizaniani, Elias Martin, Maryna Sivachenko, Afra Mashhadi, and Chirag
  Shah.
\newblock Can llms reason like humans? assessing theory of mind reasoning in
  llms for open-ended questions.
\newblock In {\em International Conference on Information and Knowledge
  Management, Proceedings}, pages 34--44. Association for Computing Machinery,
  10 2024.

\bibitem{Marjanovi2025}
Sara~Vera Marjanović, Arkil Patel, Vaibhav Adlakha, Milad Aghajohari, Parishad
  BehnamGhader, Mehar Bhatia, Aditi Khandelwal, Austin Kraft, Benno Krojer,
  Xing~Han Lù, Nicholas Meade, Dongchan Shin, Amirhossein Kazemnejad, Gaurav
  Kamath, Marius Mosbach, Karolina Stańczak, and Siva Reddy.
\newblock Deepseek-r1 thoughtology: Let's <think> about llm reasoning.
\newblock 4 2025.

\bibitem{Bashir2025}
Aneesa Bashir, Rong Peng, and Yongchang Ding.
\newblock Logic-infused knowledge graph qa: Enhancing large language models for
  specialized domains through prolog integration.
\newblock {\em Data and Knowledge Engineering}, 157:102406, 5 2025.

\bibitem{Putta2024}
Pranav Putta, Edmund Mills, Naman Garg, Sumeet Motwani, Chelsea Finn, Divyansh
  Garg, and Rafael Rafailov.
\newblock Agent q: Advanced reasoning and learning for autonomous ai agents.
\newblock {\em arXiv}, 8 2024.

\bibitem{Kolagar2024}
Zahra Kolagar and Alessandra Zarcone.
\newblock Aligning uncertainty: Leveraging llms to analyze uncertainty transfer
  in text summarization.
\newblock 03 2024.

\bibitem{Tan2023}
Yiming Tan, Dehai Min, Yu~Li, Wenbo Li, Nan Hu, Yongrui Chen, and Guilin Qi.
\newblock Can chatgpt replace traditional kbqa models? an in-depth analysis of
  the question answering performance of the gpt llm family, 2023.

\bibitem{Lu2023}
Yu-An Lu and Yu-Ting Lin.
\newblock Characterised {LLM}s affect its evaluation of summary and
  translation.
\newblock In Daniel Deutsch, Rotem Dror, Steffen Eger, Yang Gao, Christoph
  Leiter, Juri Opitz, and Andreas R{\"u}ckl{\'e}, editors, {\em Proceedings of
  the 4th Workshop on Evaluation and Comparison of NLP Systems}, pages
  184--192, Bali, Indonesia, November 2023. Association for Computational
  Linguistics.

\bibitem{Bian2024}
Tian Bian, Yifan Niu, Heng Chang, Divin Yan, Junzhou Huang, Yu~Rong, Tingyang
  Xu, Jia Li, and Hong Cheng.
\newblock Hierarchical graph latent diffusion model for conditional molecule
  generation.
\newblock In {\em International Conference on Information and Knowledge
  Management, Proceedings}, pages 130--140. Association for Computing
  Machinery, 10 2024.

\bibitem{Menis-Mastromichalakis2024}
Orfeas Menis-Mastromichalakis, Giorgos Filandrianos, Jason Liartis, Edmund
  Dervakos, and Giorgos Stamou.
\newblock Semantic prototypes: Enhancing transparency without black boxes.
\newblock 7 2024.

\bibitem{Liu2024}
Lianjun Liu, Hongli An, Pengxuan Chen, and Longxiang Ye.
\newblock A contemporary overview: Trends and applications of large language
  models on mobile devices.
\newblock 12 2024.

\bibitem{Kojima2022}
Machel Reid Yutaka~Matsuo Takeshi~Kojima, Shixiang Shane~Gu and Yusuke Iwasawa.
\newblock Large language models are zero-shot reasoners.
\newblock 2022.

\bibitem{Wei2022}
Jason Wei, Yi~Tay, Rishi Bommasani, Colin Raffel, Barret Zoph, Sebastian
  Borgeaud, Dani Yogatama, Maarten Bosma, Denny Zhou, Donald Metzler, Ed~H.
  Chi, Tatsunori Hashimoto, Oriol Vinyals, Percy Liang, Jeff Dean, and William
  Fedus.
\newblock Emergent abilities of large language models.
\newblock 2022.

\bibitem{Zhang2019}
Zhanqiu Zhang, Jianyu Cai, Yongdong Zhang, and Jie Wang.
\newblock Learning hierarchy-aware knowledge graph embeddings for link
  prediction.
\newblock 11 2019.

\bibitem{Madaan2023}
Aman Madaan, Niket Tandon, Prakhar Gupta, Skyler Hallinan, Luyu Gao, Sarah
  Wiegreffe, Uri Alon, Nouha Dziri, Shrimai Prabhumoye, Yiming Yang, Shashank
  Gupta, Bodhisattwa~Prasad Majumder, Katherine Hermann, Sean Welleck, Amir
  Yazdanbakhsh, and Peter Clark.
\newblock Self-refine: Iterative refinement with self-feedback.
\newblock 3 2023.

\bibitem{Yao2023}
Shunyu Yao, Jeffrey Zhao, Dian Yu, Nan Du, Izhak Shafran, Karthik Narasimhan,
  and Yuan Cao.
\newblock React: Synergizing reasoning and acting in language models.
\newblock {\em 11th International Conference on Learning Representations, ICLR
  2023}, pages 1--33, 2023.

\bibitem{Mishra2018}
Bhavana~Dalvi Mishra, Lifu Huang, Niket Tandon, Wen tau Yih, and Peter Clark.
\newblock Tracking state changes in procedural text: A challenge dataset and
  models for process paragraph comprehension.
\newblock 5 2018.

\bibitem{Ranaldi23}
Leonardo Ranaldi and André Freitas.
\newblock Aligning large and small language models via chain-of-thought
  reasoning.
\newblock Technical report.

\bibitem{Wang2023}
Yu~Wang, Nedim Lipka, Ryan~A. Rossi, Alexa Siu, Ruiyi Zhang, and Tyler Derr.
\newblock Knowledge graph prompting for multi-document question answering.
\newblock 8 2023.

\bibitem{Jacobson2025}
Maxwell~J. Jacobson and Yexiang Xue.
\newblock Integrating symbolic reasoning into neural generative models for
  design generation.
\newblock {\em Artificial Intelligence}, 339, 2 2025.

\bibitem{Wang2024}
Siyuan Wang, Zhongyu Wei, Jiarong Xu, Taishan Li, and Zhihao Fan.
\newblock Unifying structure reasoning and language pre-training for complex
  reasoning tasks.
\newblock {\em IEEE/ACM Transactions on Audio Speech and Language Processing},
  32:1586--1595, 2024.

\bibitem{Silver2016}
David Silver, Aja Huang, Chris~J. Maddison, Arthur Guez, Laurent Sifre, George
  Van~Den Driessche, Julian Schrittwieser, Ioannis Antonoglou, Veda
  Panneershelvam, Marc Lanctot, Sander Dieleman, Dominik Grewe, John Nham, Nal
  Kalchbrenner, Ilya Sutskever, Timothy Lillicrap, Madeleine Leach, Koray
  Kavukcuoglu, Thore Graepel, and Demis Hassabis.
\newblock Mastering the game of go with deep neural networks and tree search.
\newblock {\em Nature}, 529:484--489, 2016.

\bibitem{Park2023}
Joon~Sung Park, Joseph~C. O'Brien, Carrie~J. Cai, Meredith~Ringel Morris, Percy
  Liang, and Michael~S. Bernstein.
\newblock {\em Generative Agents: Interactive Simulacra of Human Behavior},
  volume~1.
\newblock Association for Computing Machinery, 2023.

\bibitem{Simon1970}
Herbert~A Simon and Allen Newell.
\newblock Human problem solving: The state of the theory in 1970 1.
\newblock Technical report, 1970.

\bibitem{Beranek1983}
Bolt Beranek and Newman Inc.
\newblock Structure-mapping: A theoretical framework for analogy* dedre
  gentner.
\newblock Technical report, 1983.

\bibitem{Kim_2023}
Sunnie S.~Y. Kim, Elizabeth~Anne Watkins, Olga Russakovsky, Ruth Fong, and
  Andrés Monroy-Hernández.
\newblock “help me help the ai”: Understanding how explainability can
  support human-ai interaction.
\newblock In {\em Proceedings of the 2023 CHI Conference on Human Factors in
  Computing Systems}, CHI ’23, page 1–17. ACM, April 2023.

\bibitem{Gurumoorthy2017}
Karthik~S. Gurumoorthy, Amit Dhurandhar, Guillermo Cecchi, and Charu Aggarwal.
\newblock Efficient data representation by selecting prototypes with importance
  weights.
\newblock 7 2017.

\bibitem{Peng2023}
Baolin Peng, Chunyuan Li, Pengcheng He, Michel Galley, and Jianfeng Gao.
\newblock Instruction tuning with gpt-4.
\newblock pages 1--12, 2023.

\bibitem{Chen2023}
Lingjiao Chen, Matei Zaharia, and James Zou.
\newblock Frugalgpt: How to use large language models while reducing cost and
  improving performance.
\newblock 2023.

\bibitem{BurkhardtDiss}
Daniel Burkhardt.
\newblock {\em Unleashing Autonomization: A Holistic Ontology of Data- Driven
  Solution Design realizing Business Opportunities based on Distributed Ledger
  Technology and Machine Learning}.
\newblock Steinbeis Verlag, 2025.

\bibitem{Ma2024}
Shengjie Ma, Chengjin Xu, Xuhui Jiang, Muzhi Li, Huaren Qu, Cehao Yang, Jiaxin
  Mao, and Jian Guo.
\newblock Think-on-graph 2.0: Deep and faithful large language model reasoning
  with knowledge-guided retrieval augmented generation.
\newblock 7 2024.

\bibitem{Liu2025}
Ben Liu, Jihai Zhang, Fangquan Lin, Cheng Yang, Min Peng, and Wotao Yin.
\newblock Symagent: A neural-symbolic self-learning agent framework for complex
  reasoning over knowledge graphs.
\newblock {\em arXiv}, 2022-March, 2025.

\bibitem{Mavromatis2022}
Costas Mavromatis and George Karypis.
\newblock Rearev: Adaptive reasoning for question answering over knowledge
  graphs.
\newblock 10 2022.

\bibitem{Saikh2022}
Tanik Saikh, Tirthankar Ghosal, Amish Mittal, Asif Ekbal, and Pushpak
  Bhattacharyya.
\newblock Scienceqa: a novel resource for question answering on scholarly
  articles.
\newblock {\em International Journal on Digital Libraries}, 23:289--301, 9
  2022.

\bibitem{Wu2024}
Minghao Wu, Yulin Yuan, Gholamreza Haffari, and Longyue Wang.
\newblock (perhaps) beyond human translation: Harnessing multi-agent
  collaboration for translating ultra-long literary texts.
\newblock 5 2024.

\bibitem{Gou2024}
Zhibin Gou, Zhihong Shao, Yeyun Gong, Yelong Shen, Yujiu Yang, Nan Duan, and
  Weizhu Chen.
\newblock Critic: Large language models can self-correct with tool-interactive
  critiquing.
\newblock 5 2024.

\bibitem{xie2024}
Yuxi Xie, Anirudh Goyal, Wenyue Zheng, Min-Yen Kan, Timothy~P. Lillicrap, Kenji
  Kawaguchi, and Michael Shieh.
\newblock Monte carlo tree search boosts reasoning via iterative preference
  learning, 2024.

\bibitem{Guan2024}
Saiping Guan, Jiyao Wei, Xiaolong Jin, Jiafeng Guo, and Xueqi Cheng.
\newblock Look globally and reason: Two-stage path reasoning over sparse
  knowledge graphs.
\newblock In {\em International Conference on Information and Knowledge
  Management, Proceedings}, pages 695--705. Association for Computing
  Machinery, 10 2024.

\bibitem{Xue2024}
Shangzi Xue, Zhenya Huang, Xin Lin, Jiayu Liu, Longhu Qin, Tianhuang Su,
  Haifeng Liu, and Qi~Liu.
\newblock Enhancing the completeness of rationales for multi-step question
  answering.
\newblock In {\em International Conference on Information and Knowledge
  Management, Proceedings}, pages 2753--2763. Association for Computing
  Machinery, 10 2024.

\end{thebibliography}

\end{document}